# A Three-Phase Analysis of Synergistic Effects During Co-pyrolysis of Algae and Wood for Biochar Yield Using Machine Learning

Subhadeep Chakrabarti
Department of Chemical Engineering
Vishwakarma Institute of Technology
Pune, India
subhadeep.chakrabarti21@vit.edu

Saish Shinde
Department of CSIT
Symbiosis Skills and Professional University
Pune, India
saishsh@student.sspu.ac.in

*Abstract*—Pyrolysis techniques have served to be a groundbreaking technique for effectively utilising natural and man-made biomass products like plastics, wood, crop residue, fruit peels etc. Recent advancements have shown a greater yield of essential products like biochar, bio-oil and other non-condensable gases by blending different biomasses in a certain ratio. This synergy effect of combining two pyrolytic raw materials i.e co-pyrolysis of algae and wood biomass has been systematically studied and grouped into 3 phases in this research paper- kinetic analysis of co-pyrolysis, correlation among proximate and ultimate analysis with bio-char yield and lastly grouping of different weight ratios based on biochar yield up to a certain percentage. Different ML and DL algorithms have been utilized for regression and classification techniques to give a comprehensive overview of the effect of the synergy of two different biomass materials on biochar yield.
For the first phase, the best prediction of biochar yield was obtained by using a decision tree regressor with a perfect MSE score of 0.00, followed by a gradient-boosting regressor. The second phase was analyzed using both ML and DL techniques. Within ML, SVR proved to be the most convenient model with an accuracy score of 0.972 with DNN employed for deep learning technique. Finally, for the third phase, binary classification was applied to biochar yield with and without heating rate for biochar yield percentage above and below 40%. The best technique for ML was Support Vector followed by Random forest while ANN was the most suitable Deep Learning Technique.

*Index Terms*- ANN, Biochar, Co-pyrolysis, SVM, Synergy.

## I. INTRODUCTION

Throughout the decades, with the overutilization of natural resources like coal and fossil fuels has led to their deficiency, certain efforts have been made to generate fuels by innovative technologies like incorporating biomass products like animal, plant matter, and organic waste into fuels using pyrolysis techniques.
Pyrolysis involves the thermal degradation of organic or synthetic matter in the absence of oxygen to produce green fuels. This process is faster, safer and environmentally friendlier than producing fuels by traditional techniques. The main working principle behind the pyrolysis technique is the thermochemical method to decompose the organic matter of feedstocks in an inert atmosphere at high temperatures. (generally higher than 300 degrees C). It is sometimes used as an intermediate or the beginning step in
gasification processes which is followed by partial or total oxidation of primary products such as biochar, bio-oil and biogases like $CO_2$, $H_2$, $N_2$ etc. Conditions favourable for the production of biochar are essentially lower process temperature and longer vapour residence times.

As further researches were carried out in this intriguing field of pyrolysis, it was figured out that combining different feedstocks with varying chemical compositions can enhance product yield as well as the properties of biochar. This property or effect of co-pyrolysis has been shown to reduce ash content, increase heating effect value, improve absorption rate as well as enhance pore structure. Another important effect observed with the implementation of co-pyrolysis is the synergistic interactions between different feedstocks. The synergistic effect is determined by measuring the ratios of hydrogen and carbon (H/C) as well as the oxygen to carbon (O/C) ratio of the molecules. For example, a high H/C ratio of plastic wastes works synergistically with a high O/C ratio of and low H/C ratio of lignocellulosic biomass., thereby improving the quality of the products formed.

As co-pyrolysis techniques optimize the formation of desirable products of clean sustainable fuels, leveraging certain machine learning and deep learning techniques would allow us to predict the yield of biochar, bio-oil products by carefully analyzing the kinetic and process parameters, various feedstock properties as well as adjusting our operating parameters by removing those that do not have a direct effect on our product yield. Through the synergistic integration of co-pyrolysis and machine learning, bioenergy systems can become more efficient, economically viable, and environmentally sustainable, driving the transition towards a greener and more resource-efficient future.

## II.  LITERATURE REVIEW

As co-pyrolysis is a growing field in the domain of pyrolysis technology. Several research and computational experiments have been carried out to dive deeper and blur the gap between machine learning and pyrolysis techniques.

Aessa et al. [1] demonstrated the correlation between biomass and plastics by leveraging ML models like XGBoost and SHAP explanation techniques. They were able to find a negative correlation between oxygen content to that of biochar yield, with further desirable parameters like RMSE and R2 values. To understand the theoretical aspects of pyrolysis, Bin Zhao et. Al [2] studied the effect of temperature and heating rate of pyrolysis on rapeseed stems obtained from biochar, which is a common practice in China to burn unwanted rapeseed stems, resulting in wastage and air pollution. The biochar obtained had a lower pH value and ash, with enhanced properties for tackling environmental pollution. The kinetic analysis of biochar by co-pyrolysis was elaborated by Chakraborty et. Al [4] gave a detailed study on co-pyrolysis of three biomass types- Microalgae, wastewater sludge and Cedar Wood. Activation energy and enthalpy were found to be highest for cedar wood when algae and wood were blended in the ratio of 2:1 and catalyst to biomass ratio of 2:1. [7]

XGBoost models were also implemented by Prapaporn et. al [6] for predicting bio-oil yields obtained by Shapley Additive Explanation and synthetic minority using the over-sampling technique. Faisal et. al [9] examine research on co-pyrolysis, a technique proven to create high-grade bio-oil without expensive catalysts or hydrogen. Studies show it significantly improves bio-oil quality and yield, making it a simple and cost-effective method. Co-pyrolysis even enhances the value of char and gas byproducts. The use of readily available plastic and tyre waste as additives strengthens the technique's sustainability and economic potential for converting biomass. With its ability to manage waste volume, reduce landfill dependence, and offer an alternative energy source, co-pyrolysis appears to be a promising option for many countries. [11]

## III.  MATERIALS AND METHODS

Our datasets have been methodically curated by researching several datasets from acclaimed research articles, primarily from the highly reputed Journal of Analytical and Applied Pyrolysis. To shortlist our research, we aimed to go through only those papers containing the chemical kinetic and thermodynamic properties of wood and algal biomass. Since it is a general analysis of the synergistic effect between algae and biomass. We have not taken any specific species of either biomass, but rather an amalgamation of different species by referring to the research papers. As they have practically been employed to possess impressive synergistic interaction with each other. For properties estimation, our experiments were conducted in 3 phases: kinetic analysis of biomass, where our feedstock was blended with different weight ratios and kinetic parameters observed. The second phase involved delving into finding correlations of biochar yield with proximate and ultimate analysis of our biomass ratios while the third and final phase was grouping relevant parameters from the first and second phase to group them by binary classification of bio-char yield above a certain percentage.

### A.  Phase 1 -Kinetic Analysis

Parameters considered for kinetic analysis of biomass ratio of algal and wood biomasses were activation energy, exponential factor, temperature, heating rate, algal to biomass ratio and catalyst to biomass ratio as parameters to determine the biochar yield by checking correlation with activation energy parameter. For the sake of simplicity, the ratio of weight fractions was taken up to 7 possible combinations, with two involving only individual proportions for either algae or wood biomass to compare the particular effect to that of combining each other.

The dataset was first cleaned to remove any null values that might cause miscalculations in predicting our biochar yield. On running the dataset through correlation functions on Jupyter Notebook using the activation energy, it was shown to have a good correlation with algae composition, exponential factor, and catalyst-to-biomass ratio. Hence these columns were considered for further experimentation using machine learning and deep learning algorithms. Most of the correlations did not provide a score close to 1 and it is understandable as our dataset considered was quite short and hence caused overfitting. To overcome this discrepancy, we required dataset 3-4 times its size, which was implemented in our second and third phases.

It also provides evidence to our theoretical concepts as activation energy has direct relations with exponential factor and temperature based on the Arrhenius equation :

$$\log \frac{k_2}{k_1} = \frac{E_a}{2.303R}\left[\frac{T_2 - T_1}{T_1 T_2}\right]$$

As the temperature increases, the molecules move faster and therefore collide more frequently. The molecules also carry more kinetic energy. Thus, the proportion of collisions that can overcome the activation energy for the reaction increases with temperature.

*B. Phase 2- Proximate and Ultimate Analysis*

After moving in from the kinetic analysis, it is important to determine the chemical or the proximate composition of the biomass and the effect of synergistic interactions on the same. The proximate analysis involves the determination of the different compounds present in a mixture. The ultimate analysis involves the determination of the number and types of different chemical elements present in a particular compound. Therefore, these two analysis techniques are related to each other.

The parameters for proximate analysis considered were carbon content(C%), hydrogen content (H%), oxygen content (O%) and nitrogen content (N%). For ultimate analysis, we considered volatile matter(VM), fixed carbon(FC), ash content and moisture (M) along with additional parameters that provided good relation with the first phase which are heating rate, pyrolysis temperature along residence time.
Due to overfitting caused by fewer rows in the dataset for our first phase, we have tried to extend it by adding more biomass ratios of algae to wood and testing each combination with 3 different heating rates, 10, 15 and 20 degrees Celsius/min. The dataset was extended to a total of 45 rows with 11 columns to minimize the cause of overfitting.

*C. Phase 3- Biochar Classification*

After obtaining the desired parameters that will provide a positive impact on our yield of desirable products like biochar try to predict the latter via binary classification in which the biochar yield less than 40% was labelled as 0 while that which gave more than 40% was labelled as 1. Similarly, other parameters like heating rate were labelled as low, medium and high for 10,15 and 20 values respectively and for catalyst to biomass ratio was labelled as yes or no to streamline the classification process.

Interestingly, the nitrogen content was labelled to unity for all conditions as the values were more or less close to 1 and hence could be rounded off to that value, while hydrogen content was less than 10 % for all cases and could be labelled accordingly. Finally, the blending ratios were classified into 4 sets based on different compositions of algae and wood biomass.

The third phase was experimented in two different ways, the first phase was for predicting biochar yield and the second part was for predicting heating rate. Both ML and DL algorithms were used for this case as the implementation of the same is easier for classification techniques compared to regression techniques.
For DL Techniques, certain models were trained and tested for two specific techniques – DNN and CNN(Deep Neural Network and Convolutional Neural Network) methods with optimization applied to increase the learning rate of the training model. This helps us evaluate how much the existing model has to be enhanced to deliver us as much optimum result as we desire.

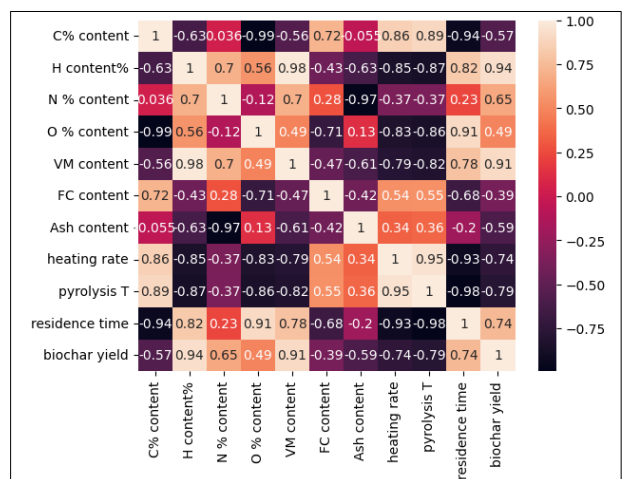

**Fig.1 Correlation heatmap obtained from the second phase**

IV. RESULTS AND DISCUSSIONS

*A. First Phase – Activation Function*

Based on the correlation heatmap obtained from the dataset, we found out that the activation function showed a good relation with algae weight fraction, exponential factor and catalyst biomass ratio. Rest other factors, however, showed negative or highly unfavourable relations. After selecting the favourable factors only from this dataset. , was run by six ML models: Linear Regression, Decision Tree Regressor, Random Forest Regressor, Ridge Regression, Support Vector Regression and Gradient Boost Regressor.

The highest accuracy or we could say the least mean square error (MSE) was observed while running the Decision Tree Regressor. with a perfect null value as well as a perfect R2 score of 1. While this may come as a surprising result. Considering the smaller number of rows, the high ensembling power of decision trees allows us to optimize the result to their desired values and also due to better hyper-tuning results.
A similar inference could be made for the Gradient Boosting model with a perfect accuracy value and an almost negligible value of MSE 1.615E-28, followed by a perfect R2 score of 1.

However, we did not run deep learning of this dataset as all the models would be highly overfitting, owing to the small length of the dataset. Hence the number of iterations was increased for the second and third phases.

B. *Proximate Analysis*

Analyzing the high overfitting owing to a small dataset in the first phase, the number of observations was increased by taking more combinations of blending ratios for wood and algae.

After running the heatmap to find correlations for predicting biochar yield, the best relation for the same was obtained with H% content, followed by volatile matter present and residence time. Another relation observed was the high correlation between pyrolysis temperature and heating rate. Hence two datasets were run for this phase: with and without heating rate as a parameter.

Without heating rate as a parameter, the following results were observed as shown in the table:

**Table 1 Scores from different ML models**

| Model Name | Accuracy Score |
|---|---|
| Linear Regression | 0.938 |
| Decision Tree | 0.939 |
| Random Forest | 0.959 |
| Ridge Regression | 0.876 |
| Support Vector | 0.972 |
| Gradient Boosting | 0.946 |
|  |  |

From the table, we observe that the highest accuracy was provided by support vector regression of 0.972. Apart from that all of the models except for ridge regression show highly favourable outcomes with an accuracy score above 90% which aligns with the fact that only highly correlated factors were included for the model implementations. The favourable values obtained from each model allowed us to apply deep learning techniques to this dataset as well.

For DL, we only tried one model which was ANN ( Artificial Neural Network). The testing size was conventionally set to 20% of training data. The optimizer used was Adam Optimizer. After setting the evaluation metrics, epochs were run for a total of 50 times to finally deliver the result of an MSE score of around 23.4225 and a loss value of 4.0771. However since DNN uses a numerical prediction, the accuracy value between 0 and 1 was not provided.

With heating rate, the most desirable results were obtained with Gradient boosting regression. With both MAE and MSE values coming as 0.658 and  0.469 respectively. For deep learning, because of the small dataset, the performance score was coming out quite low. Hence to overcome this issue, we increased the number of training examples (rows). After running DNN, the results came out to be 3.4824 loss value and MSE 19.1331 which is quite satisfactory, considering the size of the dataset.

From combining both phases, we can observe that deep learning is not that suitable for regression techniques in that particular dataset as it is quite small compared to the same techniques applied to conventional datasets involving thousands of rows with distinct values for each column. One way to overcome this is we consider carrying out a classification of the dataset which was our point of focus for the third phase.

C. *Biochar Classification*

Our classified dataset was converted to categorical variables to streamline the process. Following the methodology of the second phase, due to multiple correlations obtained in our dataset.

For DNN, we created a baseline model which even though gave a small loss value, also gave a small accuracy value.
Also for the binary classification, the model only predicted values for category 1 i.e. biochar yield above 40% and not for the category less than that. To counter this ambiguity, we increased the number of units to double its value (100 to 200), surprisingly the accuracy did not increase, but the loss value increased which was counterproductive. After checking the graphs for loss-accuracy, we noticed the loss is in a zigzag manner i.e. the values are not stable.

Hence the model 2 implementation can be safely discarded as we crosschecked the results using a diagram illustrated below. According to that, we observe the alternate peaks and troughs in the loss line . Although for the training model, the loss function does not deviate much, it is the testing model whose results we have to consider to obtain output, which in our case, gives a higher loss value than the first model.

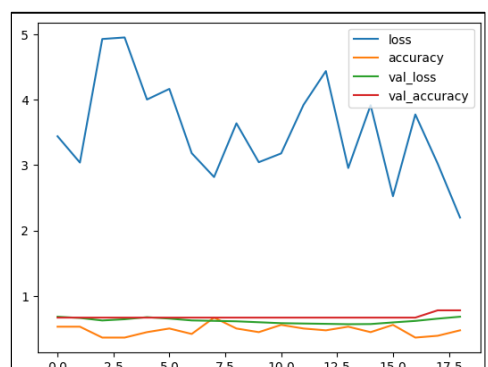

**Fig.2  Loss Accuracy Graph for Model 1**

We observed that Model 1 gave a comparatively lesser loss value than Model 2. We implemented model 1 architecture to create another model where we optimized the learning rate by decreasing it using Adam Optimizer. As expected, accuracy increased to 0.777 and loss came as 0.631 which is satisfactory (model 3). Also by plotting a confusion matrix in which both the classes gave genuine results and the dominating effect of one category was reduced.

Just to check if the learning rate could be optimized further, we tried implementing a fourth model with 150 units taken which however gave an even larger loss value of 2.146 and for the same accuracy value. Hence model number 3 was confirmed to be providing the most desirable values for accuracy and error.

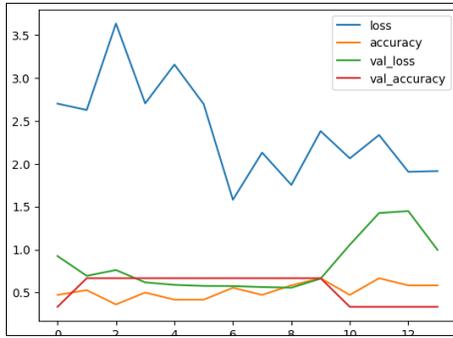

**Fig.3 Loss Accuracy Graph for Model-2 (Higher loss value)**

```
         Model1_Test  Model1_Train  Model2_Test  Model2_Train
Loss       0.697065      0.689529     0.995776      0.675938
Accuracy   0.333333      0.611111     0.333333      0.611111

         Model3_Test  Model3_Train  Model4_Test  Model4_Traint
           0.643117      0.593352     2.146178      1.258825
           0.777778      0.805556     0.333333      0.611111
```

**Fig.4 Training and Test Results for ANN and CNN**

For experimental purposes, we used CNN but got comparatively lower accuracy values and less loss (0.667 and 5.146 respectively). Surprisingly, Category 0 (undesired) was shown to be dominating over 1st category. Hence DNN model was preferred over CNN.

Moving on to ML methods, the following scores were obtained for biochar yield :

**Table 2. Results for Final phase**

| Model Used | Accuracy |
|---|---|
| LR | 0.9714 |
| Decis | 0.9333 |
| RFC | 0.9666 |
| SVC | 1.0000 |
| KNN | 1.0000 |

We used 5 models here instead of 6 as in the second phase as ridge regression did not provide the desired results in an earlier phase. Also, notice we have not used gradient boosting as we already obtained a perfect score of 1.0 in SVC and KNN. We used the KNN model for the third phase as it provides a similar output as SVC and it is easier to interpret and straightforward compared to gradient boosting.

Again, owing to the comparatively small dataset despite increasing the number of iterations, we get the perfect accuracy value of the model. Similar inferences can be considered for the KNN model which can be positively leveraged for our concerned dataset.

## V. CONCLUSIONS

To summarize all our findings from all three phases. while simpler regressions like decision trees and gradient boosting gave us higher accuracy for kinetic analysis of smaller datasets, as the iterations increased for the dataset, deep learning models could be implemented more efficiently, with optimizers like Adam utilized to increase the learning rate. ANN models provided higher accuracy for DL techniques with SVC and KNN providing perfect accuracy scores for the third phase of our training.

Co-pyrolysis is a rapidly growing technique in the field of pyrolysis techniques and leveraging machine and deep learning algorithms can help us correlate results like optimum biochar yield with synergistic effects of different biomass that will open new doors to carry about both laboratory experimentation and computation and derive newer innovative methods to generate cleaner fuels and contribute to sustainability.

.